\documentclass[letterpaper, 10 pt, conference]{ieeeconf}  %

\IEEEoverridecommandlockouts                              %
                                                          
\usepackage{bm}
\usepackage{cite}
\usepackage{amsmath,amssymb,amsfonts,mathcomp}
\usepackage{algorithmic}
\usepackage{graphicx}
\usepackage[export]{adjustbox}
\usepackage{float}
\usepackage{here}
\usepackage{textcomp}
\usepackage{algorithmic}
\usepackage{algorithm}
\usepackage{booktabs}
\usepackage{comment}
\usepackage{url}
\usepackage{hyperref}
\usepackage{makecell}

\begin{document}

\title{MirrorDrift: Actuated Mirror-Based Attacks on LiDAR SLAM}
\author{Rokuto Nagata$^{1}$, Kenji Koide$^{2}$, Kazuma Ikeda$^{1}$, Ozora Sako$^{1}$, Shion Horie$^{1}$, and Kentaro Yoshioka$^{1}$
\thanks{$^{1}$ Department of Electronics and Electrical Engineering, Keio University}
\thanks{$^{2}$ Department of Information Technology
and Human Factors, the National Institute of Advanced Industrial Science and Technology}
}

\maketitle

\begin{abstract}

LiDAR SLAM provides high-accuracy localization but is fragile to point-cloud corruption because scan matching assumes geometric consistency. Prior physical attacks on LiDAR SLAM largely rely on LiDAR spoofing via external signal injection, which requires sensor-specific timing knowledge and is increasingly mitigated by modern defense mechanisms such as timing obfuscation and injection rejection. In this work, we show that specular reflection offers an injection-free alternative and demonstrate an attack, MirrorDrift, that uses an actuated planar mirror to cause ghost points in LiDAR scans and systematically bias scan-matching correspondences.
MirrorDrift optimizes mirror placement, alignment, and actuation. In simulation, it increases the average pose error (APE) by 6.1× over random placement, degrading three SLAM systems to 2.29–3.31 m mean APE. In real-world experiments on a modern LiDAR with state-of-the-art interference mitigation, it induces localization errors of up to 6.03 m. To the best of our knowledge, this is the first successful SLAM-targeted attack against production-grade secure LiDARs.

\end{abstract}

\section{Introduction}
\label{introduction}
Accurate localization is essential for autonomous robots, and high precision is required for safe navigation and motion control in real-world environments~\cite{sato2021dirty}. LiDAR is widely used as a key sensing modality for high-precision localization because it provides 3D point clouds with long measurement range and high geometric accuracy. In particular, LiDAR SLAM commonly estimates motion by registering point clouds across consecutive frames, and its performance therefore strongly depends on the geometric fidelity of the observed point clouds.

While this reliance on geometric consistency enables accurate localization, it also creates an inherent vulnerability: pose estimation is compromised when the observed point cloud is adversarially manipulated. LiDAR spoofing, which injects or removes points via external laser illumination, has been shown to increase localization error~\cite{nagata2025slamspoof}, leading to safety risks and operational losses. However, modern LiDARs incorporate built-in interference mitigation mechanisms, such as randomized emission patterns and signal authentication, that suppress external optical interference~\cite{sato2024lidar}. Although not designed specifically for security, these mechanisms significantly increase the difficulty of signal injection attacks; prior SLAM spoofing demonstrations have been limited only to earlier-generation sensors~\cite{nagata2025slamspoof} as summarized in Table~\ref{tab_lidar_models}.

\begin{figure}[t]
\centering
\includegraphics[width=1.05\linewidth]{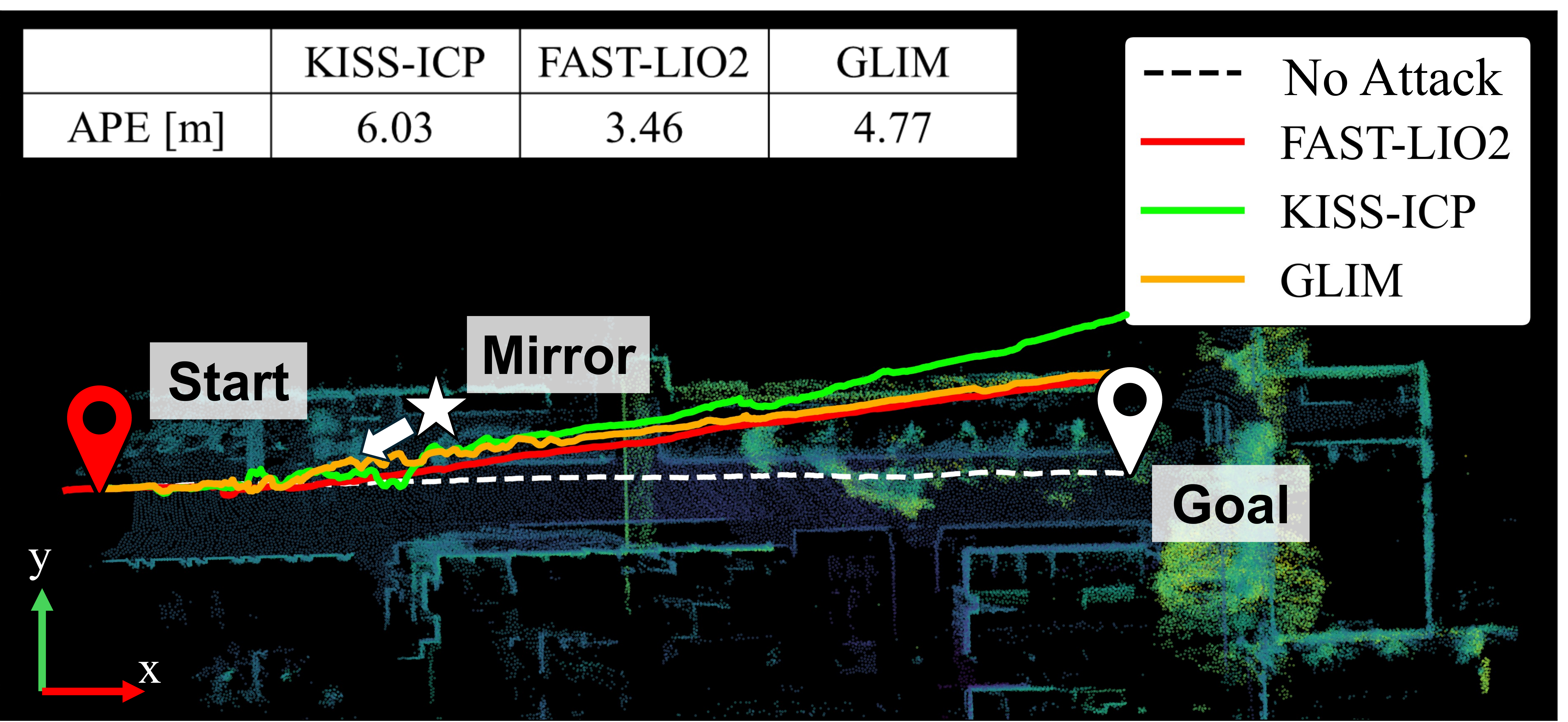}
\vspace{-0.3in}
\caption{Real-world \textit{MirrorDrift} demonstration. The reference trajectory (white dashed) and the attacked trajectories estimated by KISS-ICP~\cite{vizzo2023ral}, FAST-LIO2~\cite{fastlio2}, and GLIM~\cite{koide2024glim} (red/green/orange) show large deviations in the positive $y$-direction under attack.}
\label{fig:real_world_expetimental_result}
\end{figure}

In contrast, specular reflections from mirrors and glass generate virtual returns from the LiDAR's own emitted pulses, producing ghost points and point dropouts without any external signal injection. Because this phenomenon arises from the LiDAR's fundamental ranging process itself, it cannot be mitigated by interference rejection mechanisms designed to suppress external optical signals. Although specular surfaces are known to cause SLAM drift and loop-closure failures~\cite{liu2022integrated}, they have largely been treated as passive environmental artifacts. Whether controlled specular reflections can be systematically exploited to disrupt LiDAR SLAM remains unclear. %

Motivated by this observation, we propose MirrorDrift, an attack framework that actively exploits specular reflections using an actuated mirror placed by an adversary. MirrorDrift requires no external signal injection and continuously disrupts scan matching through reflection-induced ghost points. We further amplify the attack by optimizing mirror placement, surface alignment, and periodic yaw actuation, and validate its effectiveness through real-world experiments; it highlights specular reflection as a practical injection-free attack channel even for secure LiDARs. Our contributions are as follows:

\begin{itemize}

\item \textbf{Adversary-controlled specular-reflection attack on LiDAR SLAM:}
We demonstrate a physical attack that perturbs LiDAR SLAM localization by intentionally inducing reflection-generated ghost points using an off-the-shelf planar mirror with periodic yaw oscillation, requiring no external signal injection or specialized hardware.

\item \textbf{Design and optimization framework \textit{MirrorDrift}:}
We propose \textit{MirrorDrift}, which amplifies SLAM disruption through periodic yaw actuation and systematically optimizes mirror placement and alignment. Compared to random placement, our optimized design increases the average pose error (APE) by \textbf{6.1$\times$} on average.

\item \textbf{First real-world SLAM attack on a latest-generation secure LiDAR:}
We demonstrate up to 6.03\,m APE across three LiDAR SLAM algorithms on AT-128, a state-of-the-art automotive LiDAR with built-in interference mitigation, representing, to the best of our knowledge, the first experimental SLAM-targeted attack against such secure LiDAR systems.

\end{itemize}

\section{Related work and Background}
\label{Related work}
\subsection{LiDAR-based localization}
LiDAR SLAM provides accurate ego-localization by estimating robot motion through the registration of point clouds across consecutive frames.
In this paper, we focus on scan-matching-based LiDAR SLAM because our attack directly targets the data association and registration process. We evaluate our attack on widely used, open-source baselines that are commonly adopted in recent research and applications: KISS-ICP~\cite{vizzo2023ral} as a LiDAR-only method, and FAST-LIO2~\cite{fastlio2} and GLIM~\cite{koide2024glim} as representative LiDAR–IMU fusion methods.

\subsection{Impact of False Point Clouds on LiDAR SLAM}

\begin{table}[tb]
\centering
\caption{Comparison of attack feasibility against LiDAR models.}
\vspace{-0.1in}
\label{tab_lidar_models}
\begin{tabular}{l|cccc}
\toprule
\textbf{LiDAR model} & \textbf{VLP-32c} & \textbf{Horizon} & \textbf{HAP} & \textbf{AT-128} \\ \midrule
Appearance & 
\includegraphics[height=0.8cm, valign=m]{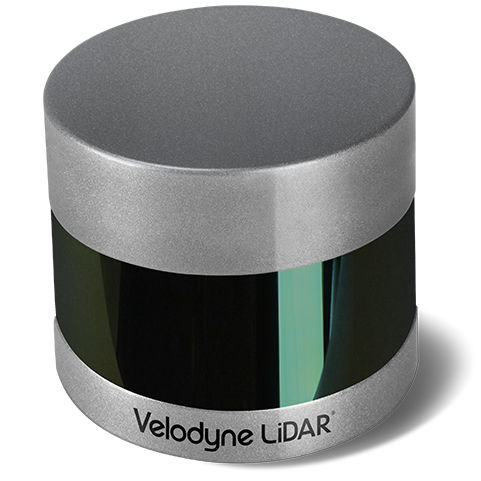} & 
\includegraphics[height=0.8cm, valign=m]{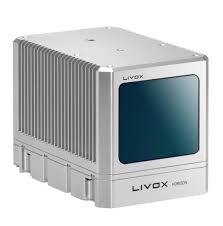} &
\includegraphics[height=1.0cm, valign=m]{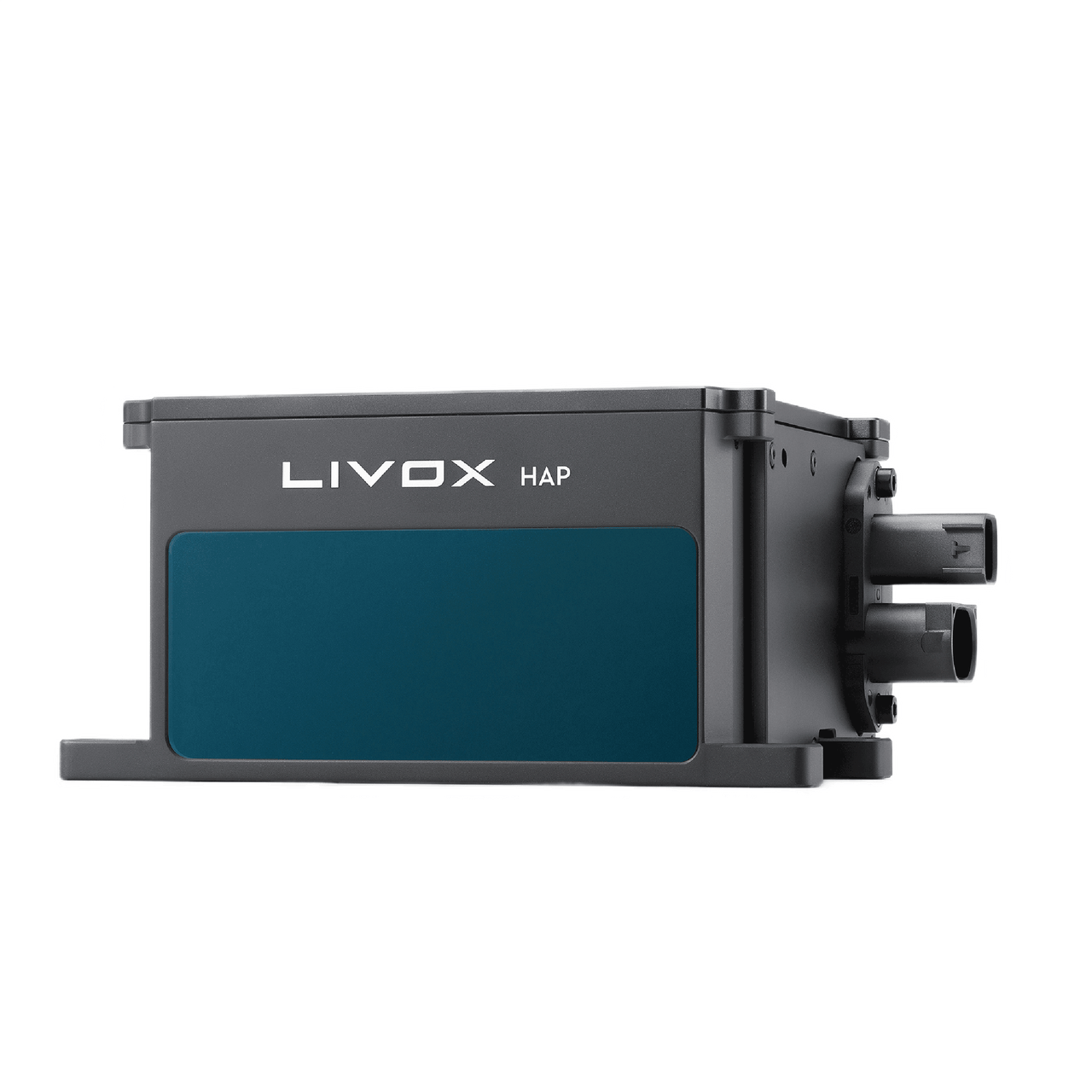} & 
\includegraphics[height=0.8cm, valign=m]{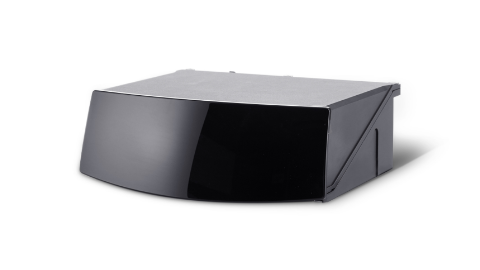} \\ \midrule
Release (Year) & 2017 & 2020 & 2020 & 2022 \\ \midrule
Channels (lines) & 32 & $\sim$48  & $\sim$64 & 128 \\ 
\makecell{Interference \\ Mitigation} & -- & \checkmark & \checkmark & \checkmark \\
\midrule
\multicolumn{5}{c}{\textit{Nagata et al.~\cite{nagata2025slamspoof}}} \\ \midrule
Real-world & \checkmark & -- & -- & -- \\
Single-frame & \checkmark & -- & -- & -- \\ \midrule
\multicolumn{5}{c}{\textit{Ours}} \\ \midrule
Real-world & -- & -- & -- & \checkmark \\
Single-frame  & \checkmark & \checkmark & \checkmark & \checkmark \\ 
\bottomrule
\multicolumn{5}{p{0.95\linewidth}}{\footnotesize
\checkmark: Experiment conducted. --: Not evaluated. Horizon and HAP employ randomized laser emission patterns.
AT-128 implements fingerprinting-based authorization.
}
\end{tabular}
\end{table}

Point-cloud corruption in LiDAR SLAM has been studied along two distinct lines: (i) non-adversarial artifacts caused by specular reflections in the environment and (ii) adversarial point-cloud manipulation created by an external attacker.

\noindent\textbf{Specular reflections in LiDAR SLAM:}
Prior studies report that mirror-like surfaces can produce reflected “virtual” structures that are observed as point clouds, leading to incorrect map updates. Such reflection-induced points can also act as noise in scan matching, causing loop-closure failures and degrading localization accuracy~\cite{yang2008dealing, 10164614, liu2022integrated, koch2015detection, tibebu2021lidar}.

\noindent\textbf{Physical attacks to LiDAR SLAM:}
In contrast, prior works ~\cite{fukunagarandom, nagata2025slamspoof} proposed LiDAR-spoofing-based attacks that inject fake points via external laser illumination, and demonstrated that it can induce large localization errors across multiple SLAM systems. 
However, injection-based attacks rely heavily on sensor-specific hardware details. For example, Chosen Pattern Injection (CPI) can induce severe localization errors~\cite{nagata2025slamspoof}. To generate an arbitrary-shaped fake point cloud at a targeted direction and range, an attacker must typically synchronize with (i) the scan pattern, (ii) beam emission timing, and (iii) the receiver time window. As analyzed in~\cite{sato2024lidar}, such synchronization becomes infeasible on recent LiDAR models that obscure timing information or incorporate pulse fingerprinting mechanisms to reject external injections. Consequently, CPI-based spoofing has not been demonstrated on modern LiDARs equipped with these protections~\cite{sato2024lidar}.

Motivated by these limitations, we propose a LiDAR SLAM attack that exploits specular reflections rather than signal injection—to generate spurious points, thereby reducing reliance on proprietary sensor implementations and injection-rejection mechanisms. In particular, we evaluate the effectiveness of our attack against a LiDAR model equipped with pulse fingerprinting.

\subsection{Attacks on LiDAR via Reflective Surfaces}
\label{mirror_attack}

Several studies have exploited reflective surfaces to manipulate LiDAR returns and affect perception. Zhu \textit{et al.}~\cite{zhu2021can} showed that highly reflective objects alter return intensity and measured range, inducing false negatives in object detection. Kobayashi \textit{et al.}~\cite{kobayashi2025invisible} proposed \textit{Shadow Hack}, which uses specular sheets to remove ground returns and create artificial shadows that trigger false positives. Yahia \textit{et al.}~\cite{yahia2025seeing} demonstrated mirror-based attacks that inject spurious points (Object Addition) or suppress returns from real objects (Object Removal).
These works focus on perception tasks and do not evaluate their impact on LiDAR SLAM localization using standard metrics such as APE or RPE. When optimization is considered, it aims to maximize detection-level effects rather than to systematically induce scan-matching failures, which are the root cause of SLAM drift.

Reflective-surface-induced ghost points have also been reported to affect SLAM stability~\cite{liu2022integrated}. However, such effects have been treated as incidental environmental artifacts rather than adversarially controlled mechanisms. To the best of our knowledge, no prior work systematically designs and actuates reflective surfaces to intentionally maximize SLAM localization drift or quantitatively evaluates such attacks under standard localization metrics.

\section{Methodology}

\begin{figure*}
    \centering
    \includegraphics[width=0.85\linewidth]{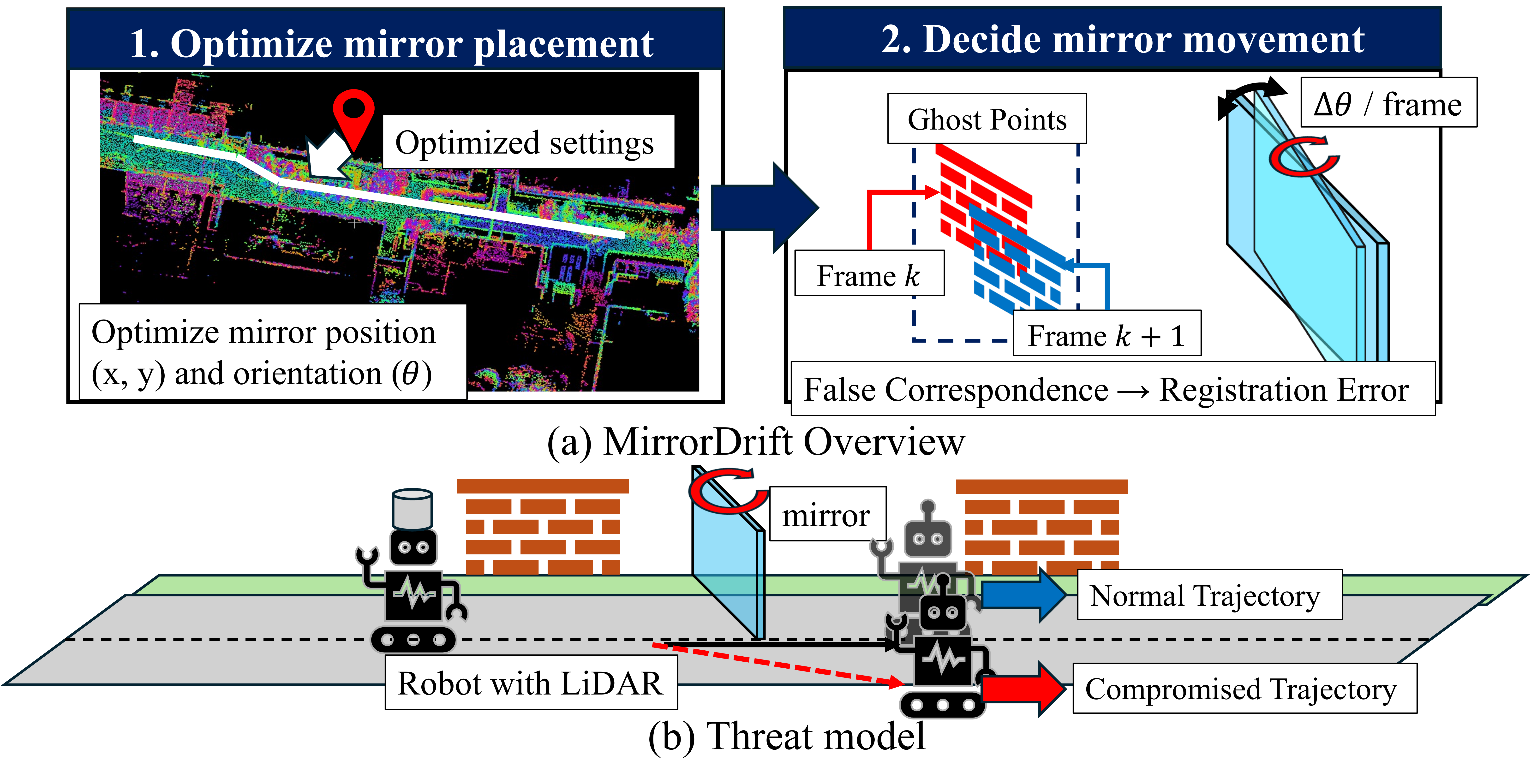}
    \vspace{-0.2in}
    \caption{(a) Overview of \textit{MirrorDrift}. We optimize the mirror placement and orientation and determine its motion pattern. The mirror position and nominal orientation are optimized by maximizing the objective function defined in \S~\ref{methodology:mirror_placement_optimize}. To induce false point-to-point correspondences across frames, the mirror is periodically oscillated in the yaw direction. (b) Threat model overview. The adversary aims to deceive the victim vehicle’s ego-localization using mirror-induced ghost points and to cause deviations from the intended route. The mirror is placed in the environment, alongside the robot’s traveling route. The victim route is assumed to be known, allowing the adversary to pre-optimize the mirror placement and orientation using \textit{MirrorDrift}.}
    \label{fig:methodology}
\end{figure*}

\label{methodology}
We propose MirrorDrift, an attack framework that exploits specular reflections to disrupt pose estimation in LiDAR SLAM. While prior real-world SLAM attacks largely rely on LiDAR spoofing via external signal injection~\cite{nagata2025slamspoof}, such injection is increasingly restricted on modern LiDARs due to built-in interference mitigation mechanisms. MirrorDrift requires no external injection: an adversary places a planar mirror to generate reflection-induced ghost points, thereby introducing a fundamentally different, injection-free attack vector.

\vspace{-0.1in}
\subsection{Threat model}
In this study, as shown in Fig.~\ref{fig:methodology}(b), we assume a setting where an attacker can place a planar mirror alongside the driving route. This corresponds to a class of physical attack scenarios in which the adversarial artifact is placed in the environment (e.g., along the roadside) to affect passing vehicles~\cite{sato2024lidar,nagata2025slamspoof}.

\noindent\textbf{Adversary capabilities and assumptions:}
\begin{itemize}
\item The victim platform is equipped with at least one 3D LiDAR sensor and performs self-localization using LiDAR SLAM.
\item The adversary is assumed to know the driving route in advance (e.g., transport robots operating on fixed routes or vehicles on scheduled services), and can optimize the mirror placement and orientation based on this information.
\item The adversary has no access to the vehicle software; the SLAM algorithm, its parameters and LiDAR model information in use are unknown to the adversary. 
\end{itemize}

\noindent\textbf{Attack objective:}
The adversary aims to inject mirror-induced reflected points (ghost points) into the LiDAR observations and thereby perturb data association in scan matching. This causes pose update errors to accumulate over time, ultimately leading to erroneous self-localization and inducing deviations from the intended trajectory.

\noindent\textbf{Success metrics:}
We primarily evaluate localization errors using APE and RPE. In addition, as a safety-oriented reference, we follow the lateral lane-departure thresholds reported by Sato \textit{et al.}~\cite{sato2021dirty}: 0.29m for urban roads and 0.74m for highways.

\subsection{Mirror simulation}
To reproduce geometric changes caused by a planar mirror, we build a specular-reflection simulation based on ray casting and a virtual sensor model. The simulation focuses on reproducing the geometry of mirror-induced ghost points and does not consider multi-bounce reflections.
To reproduce the geometric changes caused by a planar mirror, we first transform the raw point cloud $\mathcal{P}_{raw}$ into the world coordinate frame where the mirror is statically anchored. We then model two processes: (i) removing occluded points, denoted by $\mathcal{P}_{occ}$, and (ii) synthesizing mirror-reflected ghost points, denoted by $\mathcal{P}_{refl}$. The final simulated point cloud $\mathcal{P}_{sim}$ is obtained by

\vspace{-0.15in}
\begin{align}
    \mathcal{P}_{\text{sim}} = (\mathcal{P}_{\text{raw}} \setminus \mathcal{P}_{\text{occ}}) \cup \mathcal{P}_{\text{refl}} .
\end{align}

To identify the region where the mirror blocks the LiDAR line of sight, we perform ray casting from the sensor position $\bm{S} \in \mathbb{R}^3$ to each point $\bm{p} \in \mathbb{R}^3$ using the direction vector $\bm{d}=\bm{p}-\bm{S}$. 
Given the plane containing the mirror, parameterized by its center $\bm{C}\in\mathbb{R}^3$ and normal vector $\bm{n}\in\mathbb{R}^3$, the ray--plane intersection parameter $u$ is computed as
\begin{align}
    u = \frac{(\bm{C} - \bm{S}) \cdot \bm{n}}{\bm{d} \cdot \bm{n}} .
\end{align}
A point on the ray is written as $\bm{S}+u\bm{d}$, where $u=0$ corresponds to the sensor origin and $u=1$ to the measured point. Therefore, we select points whose intersection lies between $\bm{S}$ and $\bm{p}$, i.e., $0<u\leq 1$, and further check whether the intersection falls within the physical extent of the mirror. Specifically, we transform the intersection point in the world frame, $I_{\text{world}}$, into the mirror coordinate frame, $I_{\text{mirror}}$, and evaluate whether it is inside the mirror boundary. We define the set of points satisfying all these geometric constraints as the occluded point set $\mathcal{P}_{\text{occ}}$.

To generate the mirror-induced ghost points, we introduce a virtual sensor $\bm{S_v}$ that is symmetric with respect to the mirror plane. The virtual sensor position is computed as
\begin{align}
    \bm{S_v} = \bm{S} - 2 \{(\bm{S} - \bm{C}) \cdot \bm{n}\}\bm{n},
\end{align}
which enables us to simulate the physical property of specular reflection. We first check whether the sensor faces the reflective side of the mirror, i.e., $(\bm{S} - \bm{C}) \cdot \bm{n} > 0$, and extract a point set $\mathcal{P}$ whose reflection paths remain within the mirror frame. The reflected point set $\mathcal{P}_{\text{refl}}$ is then obtained by reflecting the original points across the mirror plane:
\begin{align} 
    \mathcal{P}_{\text{refl}} = \{ g(\bm{p}) \mid \bm{p} \in \mathcal{P} \} , 
    g(\bm{p}) &= \bm{p} - 2\big((\bm{p}-\bm{C})\cdot \bm{n}\big)\bm{n}.
\end{align}

This model reproduces the positions of non-existent, mirror-induced ghost points and allows us to simulate their impact on scan matching. Note that the reflection-range test that accounts for the mirror size can be implemented in the same manner as the occlusion test.

\subsection{Optimization of Mirror Placement Parameters}
\label{methodology:mirror_placement_optimize}
To maximize the impact of mirror-induced ghost points on scan matching, we design an objective function and optimize the mirror placement and orientation (alignment), as illustrated in Fig.~\ref{fig:methodology}(a).
Many scan-matching methods (e.g., ICP) estimate a rigid-body transformation that aligns two consecutive point clouds by minimizing distances between corresponding points (or local planes).
Point-cloud manipulation by a mirror-based attack device consists of two phenomena: (i) point dropout caused by occlusion by the mirror, and (ii) injection of non-existent points caused by specular reflection.
Dropout reduces the number of correct correspondences, while injected ghost points increase incorrect correspondences; consequently, the distance-minimization process converges to an erroneous pose, leading to biased motion estimates.
In iterative solvers such as Gauss--Newton ICP, pose updates are computed from residuals and their Jacobians; perturbing the correspondence set and residual distribution therefore biases the incremental pose update. Motivated by this, we use the ratios of occluded (dropped) points and reflected (injected) points in the point cloud used for registration as indicators of scan-matching disruption.

Unlike methods with explicit feature definitions [16], many recent SLAM systems [4], [6], [17] perform registration on preprocessed (e.g., voxel-downsampled) point clouds. Therefore, as an algorithm-agnostic indicator, we evaluate the occlusion and reflection ratios on the point cloud downsampled by an operator $\mathcal{D}(\cdot)$ mimicking such preprocessing.

Since SLAM typically downsamples the point cloud, for each frame $t$, we define the point ratio $\eta_{x,t}$ for $x \in \{occ, refl\}$ with respect to the downsampled full point cloud $\mathcal{D}(\mathcal{P}_{raw,t})$. Here, $\mathcal{D}(\cdot)$ denotes voxel-grid downsampling, and $N_{\mathcal{Q}}$ denotes the number of points in a point set $\mathcal{Q}$:
\begin{align}
    \eta_{x,t} = \frac{N_{\mathcal{D}(\mathcal{P}_{x,t})}}{N_{\mathcal{D}(\mathcal{P}_{raw,t})}}.
\end{align}
However, a single-frame disturbance is insufficient, as it is often corrected by subsequent observations and fails to accumulate into a persistent localization error. To capture attack persistence, we introduce exponential moving average (EMA) smoothing:
\begin{align}
    EMA_{x,t} = \alpha\eta_{x,t} + (1-\alpha)EMA_{x,t-1},
\end{align}
where we initialize $EMA_{x,0}=0$. Such biased incremental updates accumulate over time in SLAM, and the EMA formulation captures this temporal persistence rather than isolated frame-level disturbances.

Finally, we define the objective as the sum of the per-frame scores over $T$ frames and solve for the mirror position and nominal orientation $(x_{\text{mirror}},y_{\text{mirror}},\theta_{\text{mirror}})$ that maximize it (the time-varying mirror orientation is defined in the next subsection).
Since mirror-based manipulation is composed of the two effects---occlusion-induced dropout and reflection-induced injection---we treat them equally and use the sum of $EMA_{\text{occ},t}$ and $EMA_{\text{refl},t}$ as the per-frame score:
\begin{align}
    J(x_{\text{mirror}},y_{\text{mirror}},\theta_{\text{mirror}})
    = \sum_{t=1}^{T} \left( EMA_{\text{occ}, t} + EMA_{\text{refl}, t} \right).
\end{align}
The variables $(x_{\text{mirror}},y_{\text{mirror}},\theta_{\text{mirror}})$ are searched under the constraint that the attacker places the mirror while maintaining a certain distance from the target route.
Because it is generally difficult to derive analytic gradients of $J$, we treat the problem as black-box optimization and maximize $J$ via Bayesian optimization.

\subsection{Mirror Motion Pattern}
\label{methodology_mirror_swing}

To systematically disrupt scan matching, we periodically oscillate the mirror yaw angle, as illustrated in Fig.~\ref{fig:methodology}(a), so that the mirror-induced ghost points change dynamically, independently of the robot motion.
When non-existent points caused by specular reflection remain as inliers during data association, the estimated transformation is biased to align with those points. 
By actively oscillating the mirror, the attacker continuously induces unnatural inter-frame displacements of ghost points, forcing sustained registration errors. 
However, to ensure these active ghost points are not rejected during matching, the mirror actuation must be bounded.

Many scan-matching methods employ a maximum correspondence distance threshold $M$ to reject outliers~\cite{segal2009generalized}; correspondences whose inter-frame displacement exceeds $M$ are discarded as outliers.
Therefore, we design $\Delta \theta$ so that the inter-frame displacement of ghost points stays within the correspondence distance threshold $M$.

\begin{figure}
    \centering
    \includegraphics[width=1.0\linewidth]{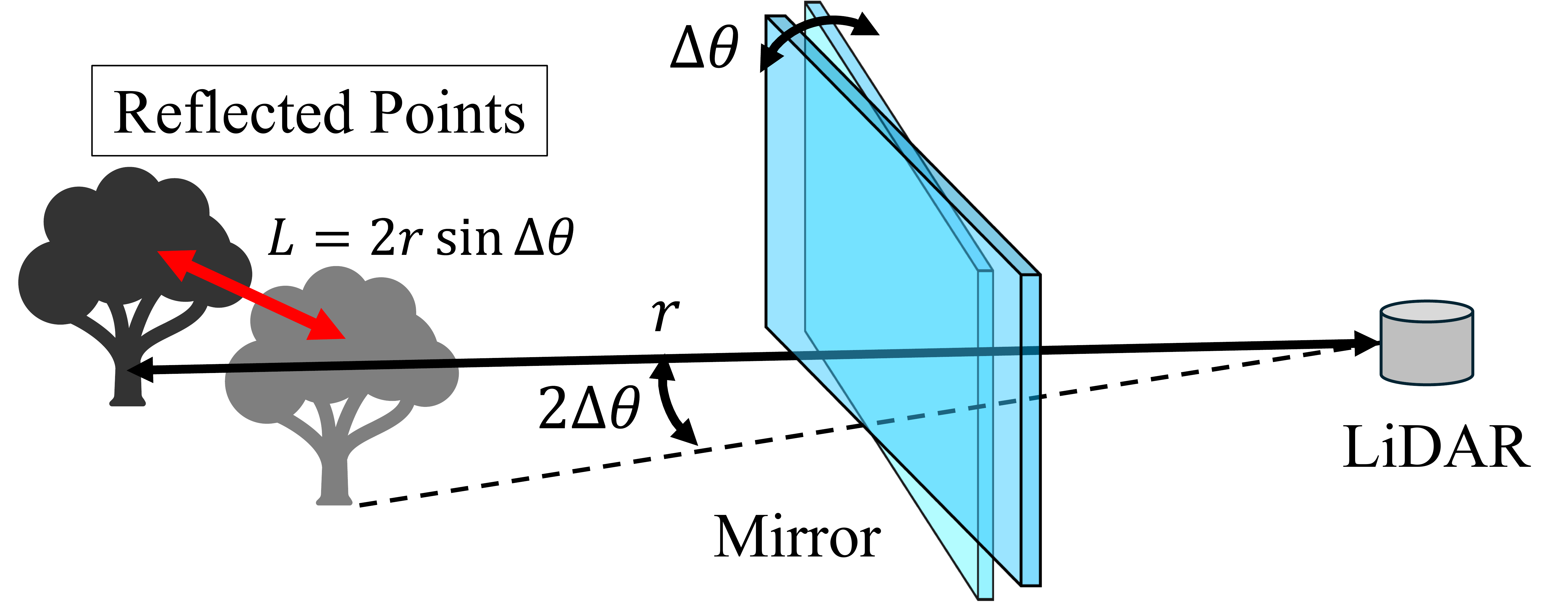}
    \vspace{-0.25in}
    \caption{Mirror-induced ghost point behavior under mirror angle changes. When the mirror rotates by $\Delta\theta$, the reflection direction changes by $2\Delta\theta$ according to the law of reflection. As a result, a ghost point at range $r$ undergoes an inter-frame displacement of $2r\sin(\Delta\theta)$.}
    \label{fig:mirror_angle_design}
\end{figure}

As illustrated in Fig.~\ref{fig:mirror_angle_design}, when the mirror yaws by $\Delta\theta$ between consecutive frames, a ghost point at range $r$ exhibits an inter-frame displacement
\begin{align}
    L_v = 2r\sin(\Delta\theta).
\end{align}
Let $r_{\max}$ denote the maximum range of the ghost point cloud. The displacement of the farthest ghost point is then $L_{v,\max}=2r_{\max}\sin(\Delta\theta)$.
If $L_{v,\max}$ exceeds the threshold $M$, the ghost points are rejected as outliers, degrading the attack effectiveness. Hence, the condition $L_{v,\max} \leq M$ dictates the maximum permissible angular step per frame for the attacker, yielding
\begin{align}
    \Delta \theta \leq \arcsin\!\left(\frac{M}{2r_{\text{max}}}\right).
\end{align}
In implementation, given the LiDAR scan period $\Delta t$, we set the mirror angular velocity $\omega$ such that $\omega \leq \Delta\theta / \Delta t$, driving the mirror yaw angle periodically to satisfy this constraint.

\subsection{Attack device design}
\label{methodology_attack_device}

\begin{figure}[tb]
\centering
\includegraphics[width=1.0\linewidth]{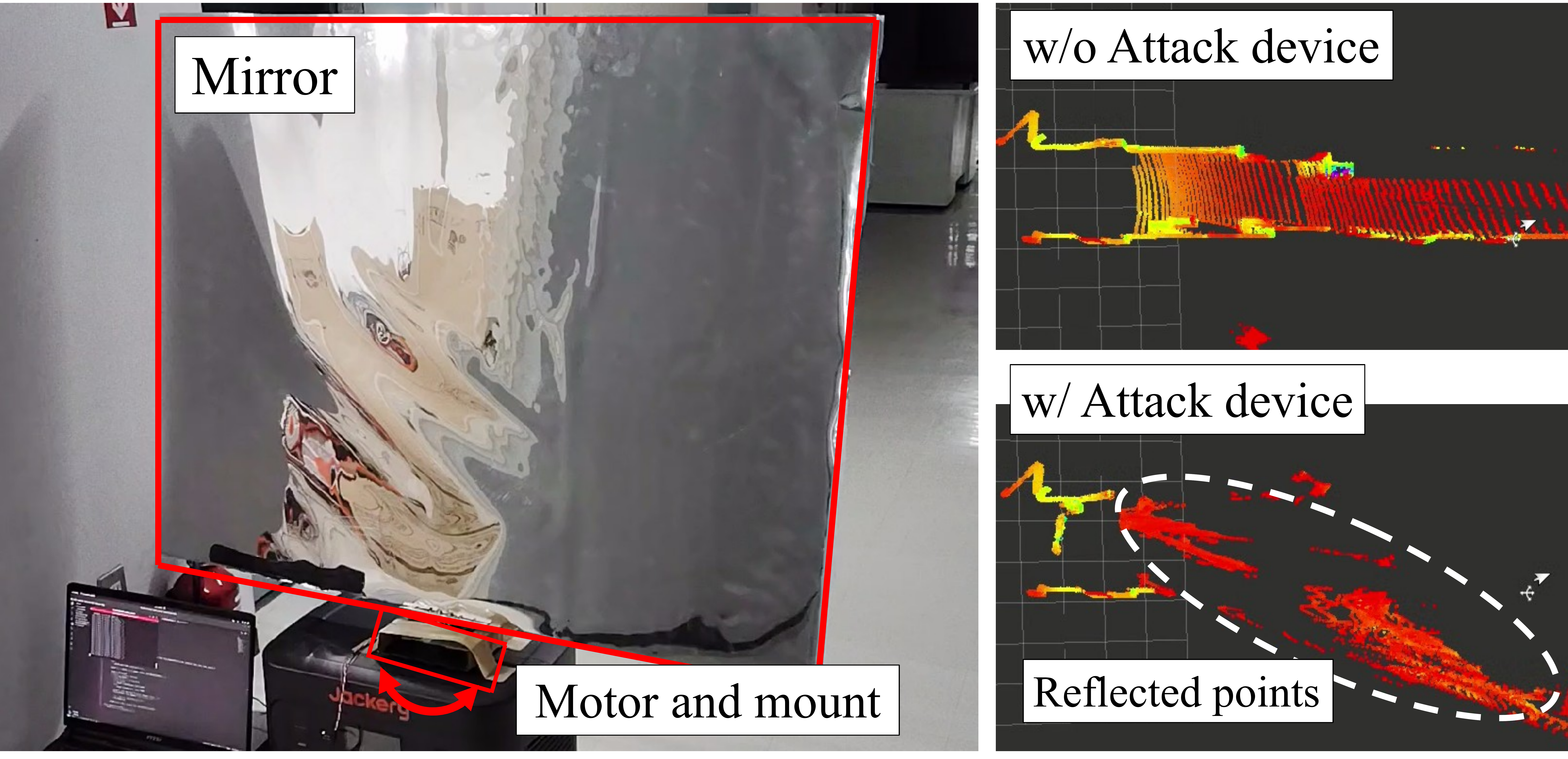}
\vspace{-0.3in}
\caption{Attack device and its effect on LiDAR measurements. (Left) Implemented mirror-actuation device consisting of a planar mirror mounted on a fixture and driven by a Dynamixel MX-28 servo motor to vary the mirror yaw angle over time. (Right) Comparison of LiDAR point clouds captured without the mirror (w/o attack device) and with the mirror present and actuated (w/ attack device).}
\label{fig:attck_device}
\end{figure}

We design and implement an attack device to realize \textit{MirrorDrift} proposed in \S ~\ref{methodology:mirror_placement_optimize} and \S ~\ref{methodology_mirror_swing}. The device actuates a planar mirror so that its yaw angle varies over time. The device consists of a planar mirror, a servo motor (Dynamixel MX-28), and a mounting fixture. Fig.~\ref{fig:attck_device} shows the implemented device and example LiDAR observations with mirror-induced ghost points.

We conduct an indoor feasibility test to confirm that the attack device can influence LiDAR observations via specular reflections. We perform this test on all four LiDAR models listed in Table~\ref{tab_lidar_models}. In each case, we collect LiDAR scans while actuating the mirror and verify that mirror-induced ghost points are injected into the observations.

\section{Evaluation}

\subsection{Validation of Mirror Simulation}
We validate our mirror-reflection simulator using real LiDAR measurements. We fixed a VLP-32c and placed a planar mirror (1.0\,m $\times$ 0.4\,m) vertically on the floor with its center aligned to the LiDAR height. We first captured a reference point cloud with the mirror present, $\mathcal{P}_{\text{ref}}$, and then removed only the mirror and captured a background point cloud. Applying our simulator to the background yields a simulated point cloud $\mathcal{P}_{\text{sim}}$.

We quantify the geometric fidelity using the Chamfer Distance~\cite{fan2017point}:
\begin{align}
d_{\text{CD}}(\mathcal{P}_{\text{ref}},\mathcal{P}_{\text{sim}})
&= \frac{1}{N_{\mathcal{P}_{\text{ref}}}}
\sum_{x \in \mathcal{P}_{\text{ref}}} \min_{y \in \mathcal{P}_{\text{sim}}} \|x - y\|_2^2 \nonumber\\
&\quad + \frac{1}{N_{\mathcal{P}_{\text{sim}}}}
\sum_{y \in \mathcal{P}_{\text{sim}}} \min_{x \in \mathcal{P}_{\text{ref}}} \|x - y\|_2^2 .
\end{align}

\begin{figure}
    \centering
    \includegraphics[width=1.0\linewidth]{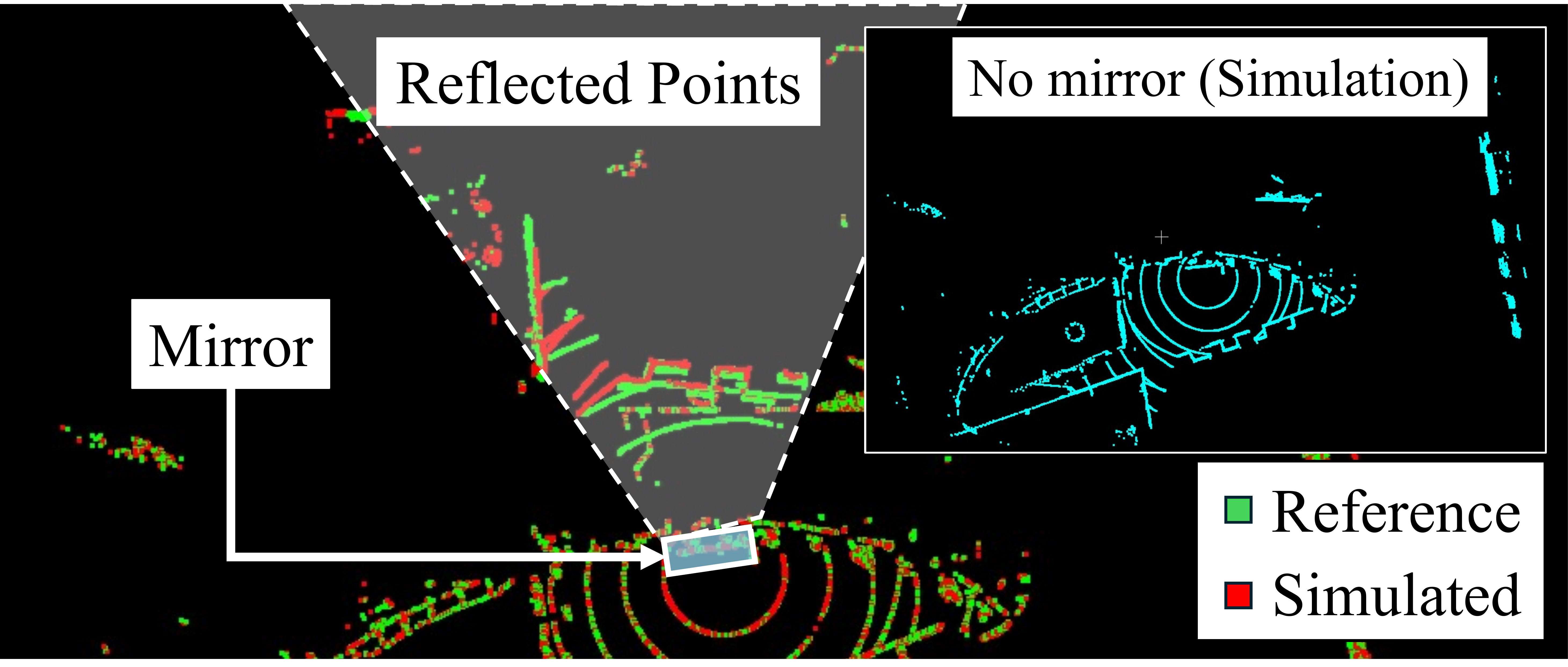}
    \vspace{-0.2in}
    \caption{Mirror simulation results. The reference point cloud is plotted in yellow-green, and the simulated point cloud is overlaid in red. The two point clouds closely overlap, yielding a very small Chamfer Distance of $0.028176$\,m.}
    \label{fig:mirror_mirror_simulation_result}
\end{figure}

\noindent\textbf{Results:}
Fig.~\ref{fig:mirror_mirror_simulation_result} shows a close overlap between $\mathcal{P}_{\text{ref}}$ and $\mathcal{P}_{\text{sim}}$, and the Chamfer Distance is $d_{\text{CD}}=0.028176$\,m, indicating that our simulator accurately reproduces the geometry of mirror-induced ghost points.

\subsection{Definition of Experimental Conditions}
\label{eval_standard_settings}
Unless otherwise stated, we use the default settings in Table~\ref{tab:default_settings}.

\noindent\textbf{Sensor setting.}
We restrict the VLP-32c horizontal FoV to the front-facing $180^\circ$ for consistency with the LiDARs used in \S ~\ref{eval_real_world_study}; the reference trajectory is generated under the same FoV.

\noindent\textbf{Mirror setting.}
We use a $1.8\,\mathrm{m}\times0.9\,\mathrm{m}$ planar mirror and drive its yaw periodically at $7.0~\mathrm{deg/s}$. The nearest distance between the mirror and the target trajectory is set to $1.5$\,m.

\begin{figure}
    \centering
    \includegraphics[width=1.0\linewidth]{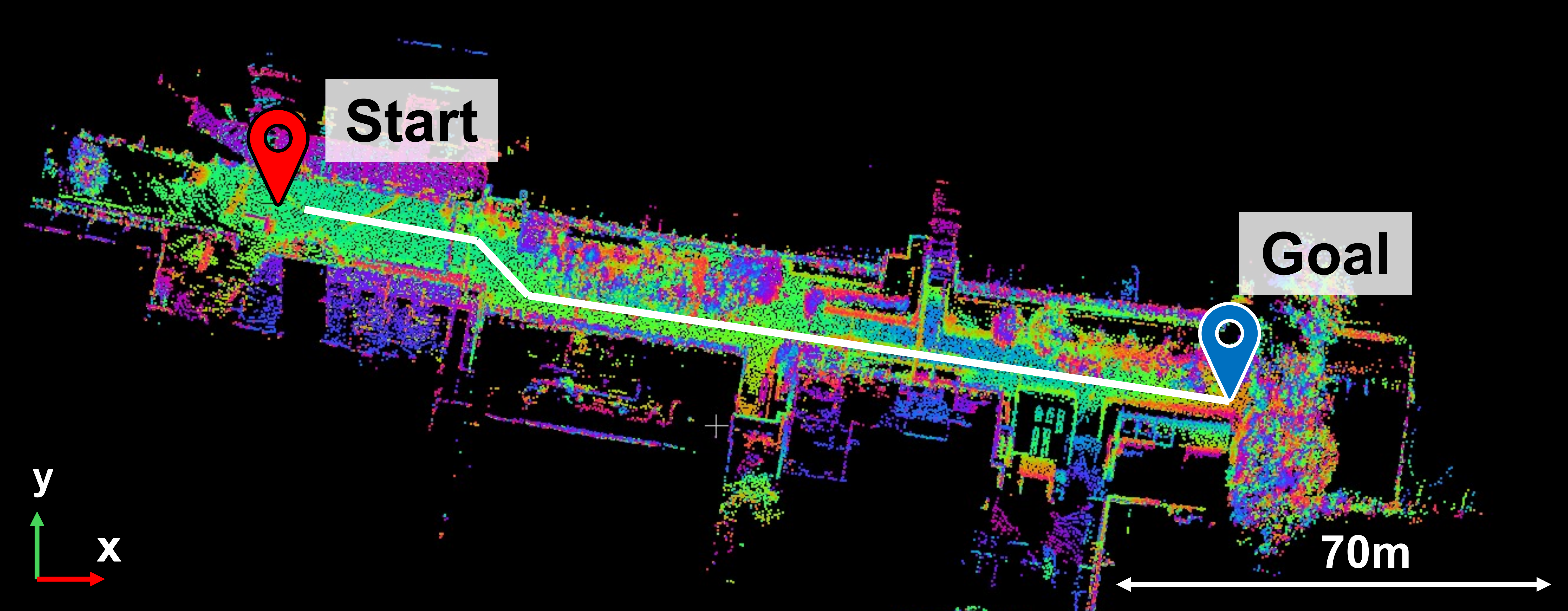}
    \vspace{-0.25in}
    \caption{Experimental environment used in our evaluation. The Start location in the upper-left is set as the origin, and the platform moves toward the Goal location in the lower-right, proceeding in the positive $x$ and negative $y$ directions.}
    \label{fig:experiment_environment}
\end{figure}

\noindent\textbf{Environment and SLAM.}
We use the route shown in Fig.~\ref{fig:experiment_environment} and evaluate KISS-ICP~\cite{vizzo2023ral}, FAST-LIO2~\cite{fastlio2}, and GLIM~\cite{koide2024glim} with default parameters unless noted.

\noindent\textbf{Metrics.}
We report APE(RMSE)~\cite{sturm2012benchmark} and maximum heading error (i.e., yaw angle error) by comparing the estimated trajectory to the reference (mirror-absent) trajectory.

\begin{table}[t]
\centering
\caption{Default experimental settings (unless otherwise stated).}
\label{tab:default_settings}
\begin{tabular}{l l}
\toprule
LiDAR / IMU & Velodyne VLP-32c / WIT Motion WT901C \\
LiDAR horizontal FoV & $180^\circ$ (front-facing) \\
Mirror size & $1.8\,\mathrm{m}\times0.9\,\mathrm{m}$ \\
Mirror yaw motion & periodic swing, $7.0~\mathrm{deg/s}$ \\
Environment / route & Fig.~\ref{fig:experiment_environment} (Start$\rightarrow$Goal) \\
SLAM methods & KISS-ICP, FAST-LIO2, GLIM \\
SLAM parameters & default (changes are explicitly stated) \\
Metrics & APE and Heading error \\
\bottomrule
\end{tabular}

\end{table}

\subsection{Proposed Method Performance Validation}
\label{eval_performance_validation}

\noindent\textbf{Experimental settings:}
To evaluate the importance of the proposed mirror placement optimization, we conducted comparative simulations between the attack results obtained with the optimized placement and two baselines: (1) no mirror and (2) randomly placed mirrors.
For the random-placement baseline, we fixed the mirror--trajectory distance to be the same as that of the optimized placement.
We performed Bayesian optimization for 300 epochs using Optuna~\cite{akiba2019optuna} with the EMA coefficient set to $\alpha=0.3$, and obtained the optimized parameters.
All experiments follow the default settings defined in \S~\ref{eval_standard_settings}.
For each SLAM algorithm, we ran the simulation 20 times and report the mean and standard deviation of APE and Heading error.

\begin{table*}[tb]
\centering
\caption{Simulation results. Unit: meter / degree.}
\vspace{-0.1in}
\label{tab_simulation_result}
\begin{tabular}{c|cccccc}
\toprule
SLAM algorithms  & APE (no mirror) & APE (random) & APE (Optimized, ours) & Head. (no mirror) & Head. (random) & Head. (Optimized, ours) \\ \midrule
KISS-ICP~\cite{vizzo2023ral} & $0.064\pm0.080$ & $0.429\pm0.288$ & \textbf{3.305$\pm$0.047} & $0.277\pm0.346$ & $0.237\pm0.166$ & \textbf{9.700$\pm$0.516} \\
FAST-LIO2~\cite{fastlio2} & $0.198\pm0.159$ & $0.425\pm0.235$ & \textbf{2.292$\pm$0.019} & $0.710\pm0.240$ & $2.387\pm1.686$ & \textbf{6.462$\pm$0.078} \\
GLIM~\cite{koide2024glim} & $0.213\pm0.146$ & $0.551\pm0.737$ & \textbf{2.900$\pm$0.141} & $1.454\pm0.626$ & $2.960\pm4.150$ & \textbf{8.081$\pm$0.784} \\ 
\bottomrule
\end{tabular}
\end{table*}

\noindent\textbf{Results:}
As shown in Table~\ref{tab_simulation_result}, the optimized mirror placement induces the largest localization errors across all three SLAM algorithms, demonstrating that the attack is effective without relying on a specific SLAM implementation. Quantitatively, the optimized design increases the average pose error (APE) by 25.6$\times$ compared to the no-mirror baseline and by 6.1$\times$ compared to random placement, on average. These results suggest that the proposed placement optimization effectively identifies mirror configurations that exploit vulnerabilities in scan matching and amplify pose drift.

\vspace{-0.099in}
\subsection{Ablation study}
\label{eval_attack_validation}
We validate the underlying mechanism of the proposed attack by conducting an ablation study that separates the contributions of occlusion and reflection.

\noindent\textbf{Experimental settings:}
We examine whether localization errors stem from occlusion-driven point dropout or ghost-point injection by comparing three conditions: (a) occlusion + reflection, (b) occlusion only, and (c) reflection only.

\noindent\textbf{Results:}
As shown in Table~\ref{tab_ablation_study}, occlusion alone increases the error by only a few centimeters over the no-mirror baseline, whereas reflection alone induces much larger errors and explains most of the full-condition degradation. These results indicate that specular-reflection-induced ghost points dominate the proposed attack.

\begin{table}[tb]
\centering
\caption{Ablation study results. Metric is APE.(simulation)}
\vspace{-0.1in}

\label{tab_ablation_study}
\begin{tabular}{c|ccc}
\toprule
SLAM algorithms  & Occl.+Refl. & Occl. & Refl. \\ \midrule
KISS-ICP~\cite{vizzo2023ral} & \textbf{3.305$\pm$0.047}         & $0.188\pm0.022$          & $3.259\pm0.026$    \\
FAST-LIO2~\cite{fastlio2} & \textbf{2.292$\pm$0.019}          & $0.237\pm0.008$         & $2.237\pm0.220$      \\
GLIM~\cite{koide2024glim} & \textbf{2.900$\pm$0.141}          & $0.128\pm0.045$          & $2.292\pm0.617$      \\ 
\bottomrule
\end{tabular}
\end{table}

\subsection{Robustness Validation: Distance Constraint and Placement Errors}
\label{eval_robustness_validation}
In this subsection, we evaluate the robustness of \textit{MirrorDrift} under realistic conditions where an attacker cannot place the mirror exactly as optimized. Specifically, we consider two factors: (i) constraints on the mirror--trajectory distance and (ii) errors in placement position and orientation. 

\subsubsection{Mirror--trajectory distance validation.} 

\noindent\textbf{Experimental settings:}
To clarify the distance constraint inherent to a placed-in-environment attack, we evaluate how the nearest distance between the mirror and the driving trajectory affects the attack effectiveness.
Using the optimized placement (nearest distance $1.5\,\mathrm{m}$) as the reference, we translate the mirror to vary the nearest distance from $1.5$ to $4.0\,\mathrm{m}$ in increments of $0.5\,\mathrm{m}$.
To ensure consistency with the baseline conditions, all other parameters are fixed to the default settings.
For each condition, we conduct 10 trials and report the mean and standard deviation of APE.

\begin{figure}[t]
\centering
\includegraphics[width=1.0\linewidth]{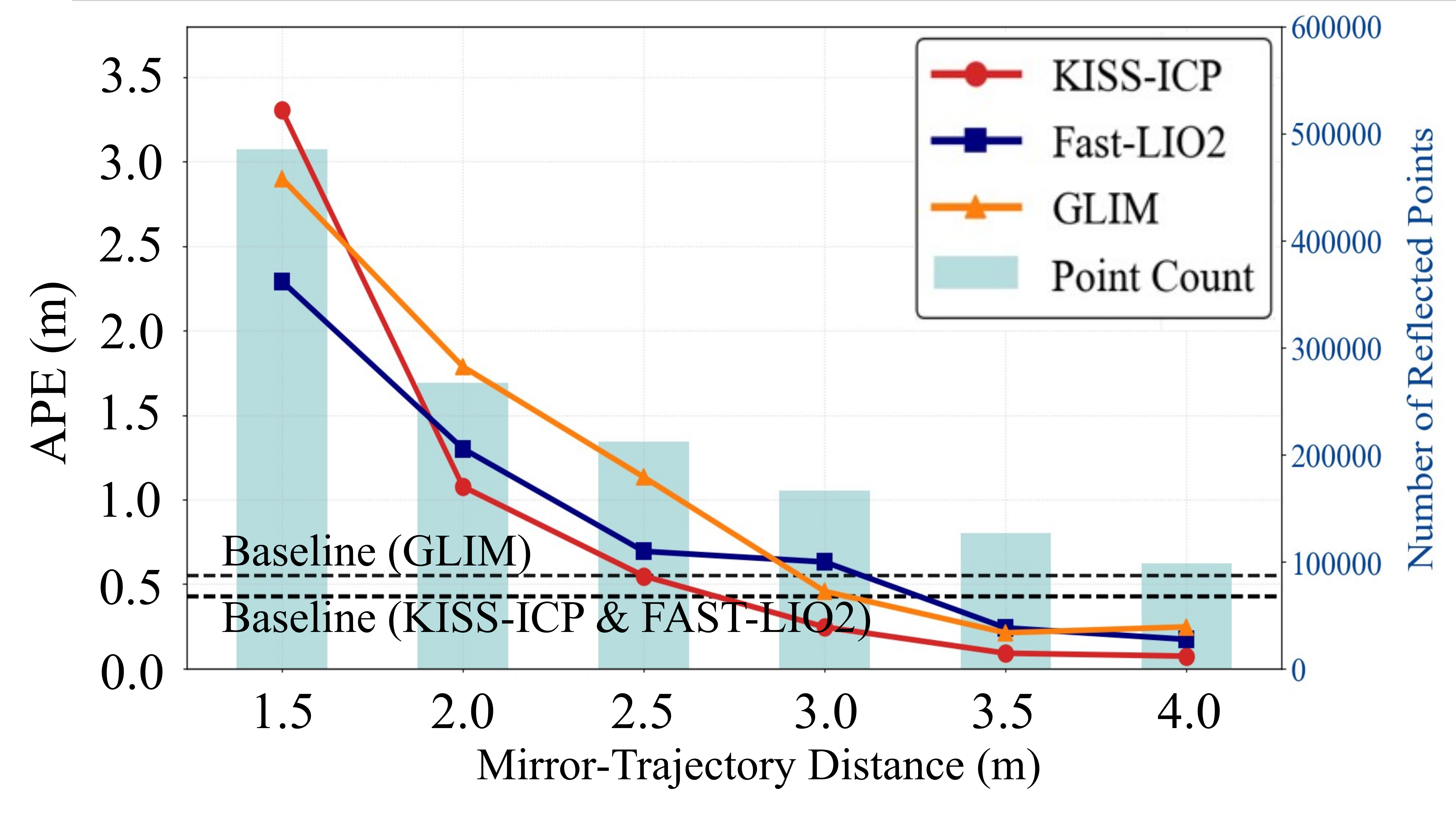}
\vspace{-0.3in}
\caption{Relationship between the mirror--trajectory distance and attack effectiveness (APE). As the distance between the mirror and the trajectory increases, the number of mirror-induced ghost points and the resulting APE decrease.}
\label{fig:mirror_traj_distance}
\end{figure}

\noindent\textbf{Results:}
As shown in Fig.~\ref{fig:mirror_traj_distance}, increasing the mirror--trajectory distance reduces the amount of observable ghost points, leading to lower APE across all SLAM algorithms.
However, when the nearest distance is $3.0\,\mathrm{m}$ or less, the optimized placement maintains substantially higher attack effectiveness than the random-placement baseline.
These results show that the mirror-trajectory distance strongly governs the quantity of mirror-induced ghost points and is a primary factor in attack effectiveness.

\subsubsection{Placement error validation} 
\noindent\textbf{Experimental settings:}
In real-world settings, precisely placing the mirror as optimized is not straightforward. We therefore investigate how errors in placement position and orientation affect attack performance. Starting from the optimized placement $(x^\star, y^\star, \theta^\star)$, we independently add position errors $\Delta x, \Delta y \sim \mathcal{U}(-0.5, 0.5)$ [m] and an angular error $\Delta\theta \sim \mathcal{U}(-5^\circ, 5^\circ)$, and randomly sample 100 perturbed placements. We evaluate APE with respect to the reference trajectory under the default settings in \S ~\ref{eval_standard_settings}. For comparison, we also evaluate a random-placement baseline with the same mirror–trajectory distance.

\begin{figure}[t]
    \centering
    \includegraphics[width=1.0\linewidth]{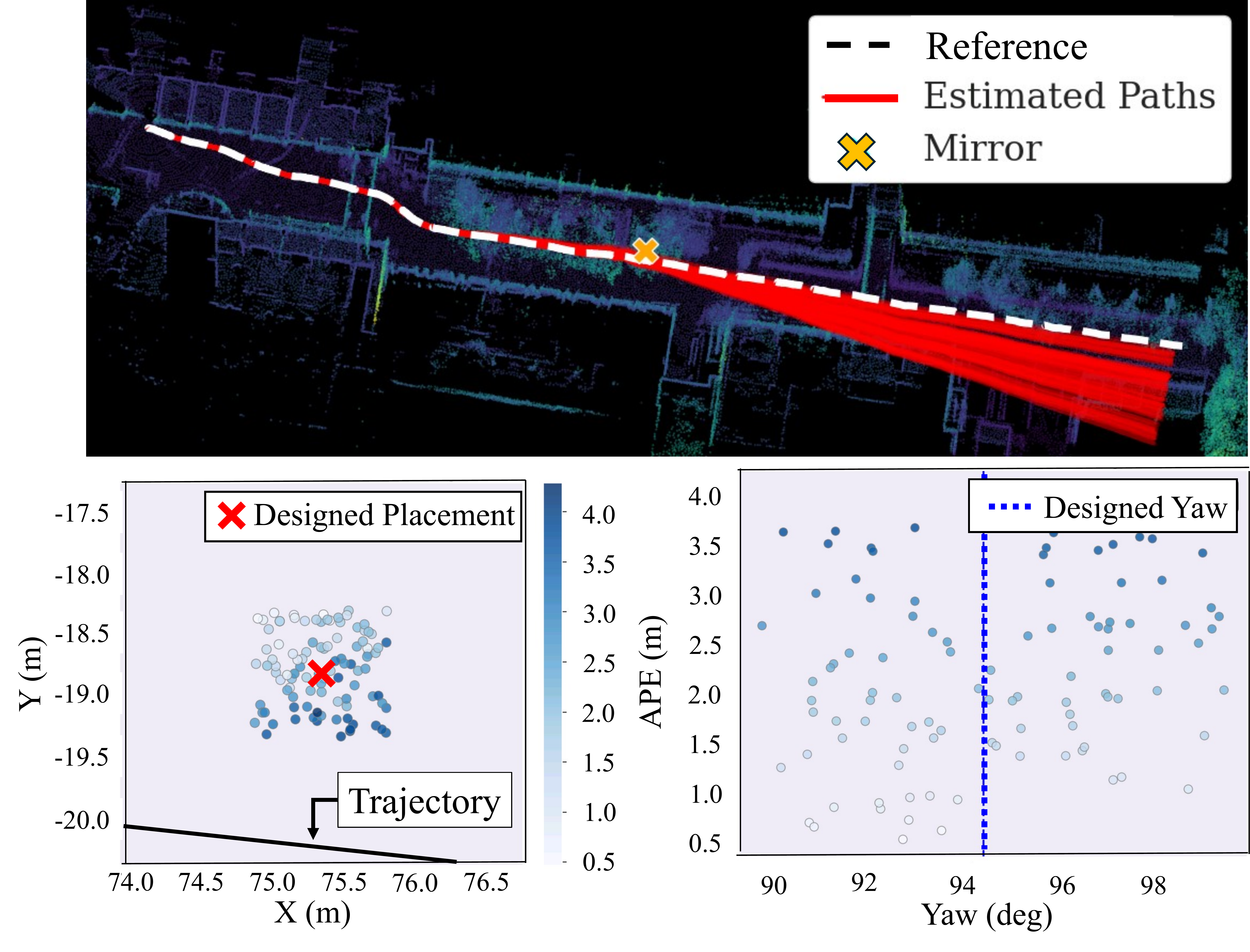}
    \vspace{-0.2in}
    \caption{(Upper) Estimated trajectories across 100 trials with stochastic placement noise ($\pm 0.5$ m for $x, y$ and $\pm 5^\circ$ for yaw). (Lower)APE heatmap over mirror pose $(x, y, \text{yaw})$(FAST-LIO2). Sustained errors across the noise range demonstrate the attack's robustness, showing that coarse placement within the LiDAR's FoV suffices without cm-level precision.}
    \label{fig:placement_error}
    \end{figure}

\noindent\textbf{Results:}
As shown in Fig.~\ref{fig:placement_error}, even with placement errors ($\pm0.5\,\mathrm{m}$ and $\pm5^\circ$), \textit{MirrorDrift} induces larger APE than the random-placement baseline.
We also observe that APE varies relatively little with yaw-angle errors, whereas it changes substantially with variations in the placement position, particularly along the $y$ direction. 
This is because the nearest mirror--trajectory distance tends to dominate the amount of mirror-induced ghost points, and distance changes strongly affect the attack effectiveness.

\subsection{Real-World Experiments}
\label{eval_real_world_study}

\begin{figure}[t]
\centering
\includegraphics[width=0.8\linewidth]{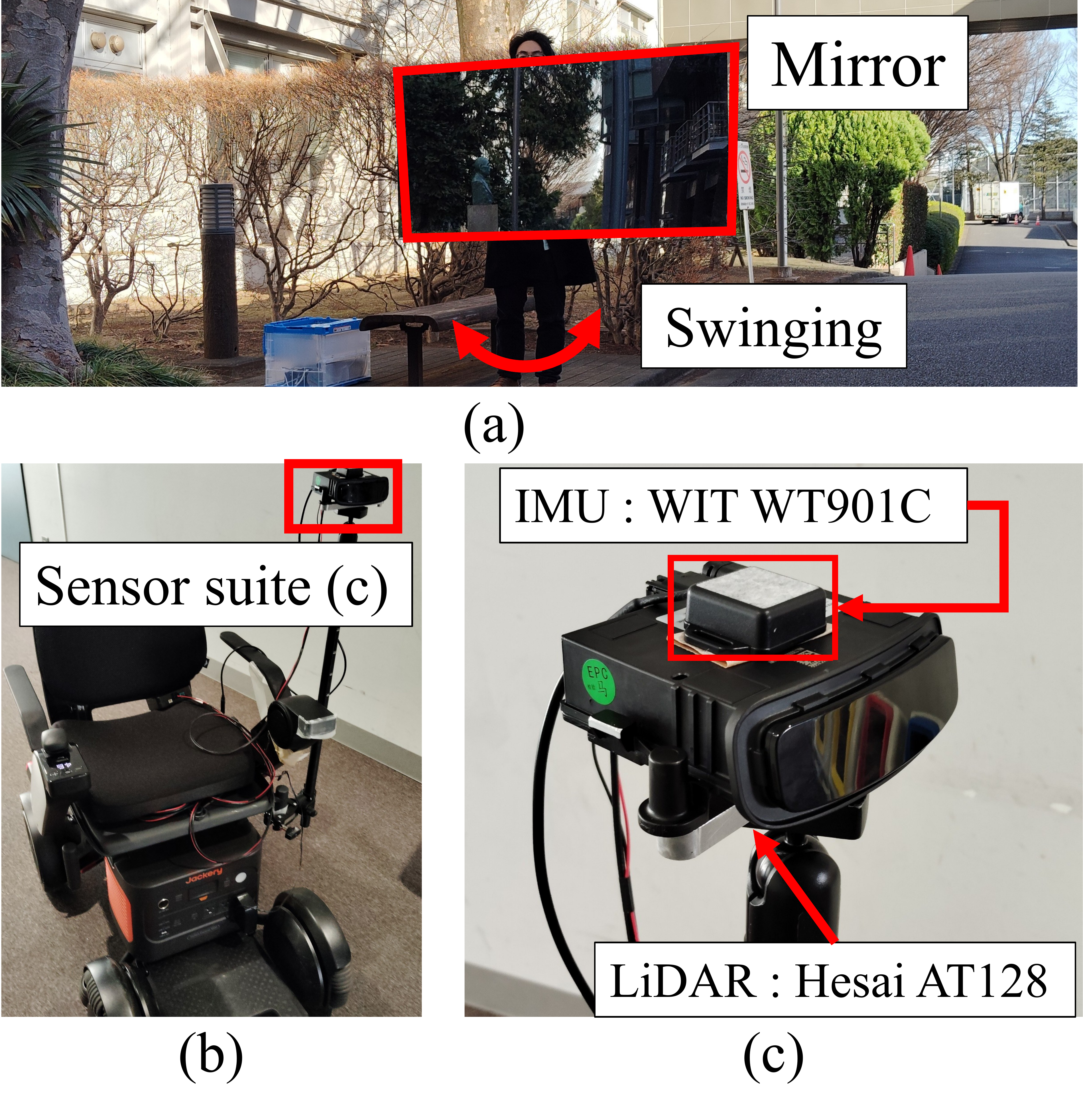}
\vspace{-0.2in}
\caption{Real-world experimental setup. (a) Mirror in the experimental environment (planar mirror, $1.6,\mathrm{m}\times0.8,\mathrm{m}$), manually oscillated in yaw. (b) Victim vehicle (electric wheelchair) with sensors. (c) Sensor system: Hesai AT128 LiDAR and an external IMU (WT-901C).}
\label{fig:real_world_expetimental_setting}
\end{figure}

Finally, we demonstrate that the proposed method can constitute a feasible physical attack exploiting specular reflections even against modern LiDAR sensors equipped with countermeasures such as protection against external signal injection.\\
\noindent\textbf{Experimental settings:}
We conducted the real-world validation using the Hesai AT128. The AT128 can identify self-emitted pulses via pulse fingerprinting and is one of the current high-security LiDAR sensors with robustness against external signal injection.

In the experiment, we mounted the sensors on an electric wheelchair as shown in Fig.~\ref{fig:real_world_expetimental_setting}(b) and collected data along a predefined outdoor course.
We first collected a run without the mirror, where the robot successfully localizes, and used it as the reference trajectory.
The reference trajectory was constructed by transforming RTK-GNSS positioning results into the robot coordinate frame.
As shown in Fig.~\ref{fig:real_world_expetimental_setting}(c), the LiDAR and IMU were rigidly attached to the vehicle body.

During the attack, we used a commercially available full-length mirror ($1.6\,\mathrm{m}\times0.8\,\mathrm{m}$) as the attack device, and adopted the placement and reflective-surface orientation optimized by the proposed \textit{MirrorDrift} framework. Although we implemented a motorized mirror device (\S ~\ref{methodology_attack_device}), in the outdoor experiment we actuated the mirror manually due to weather conditions (e.g., wind) that made stable device operation difficult. Attack success is evaluated by comparing the SLAM-estimated trajectory under attack with the reference trajectory.

\noindent\textbf{Results:}
As shown in Fig.~\ref{fig:real_world_expetimental_result}, the real-world experiments exhibit significant localization drift for all three SLAM algorithms evaluated in this study.
Table~\ref{tab_realworld_result} reports the APE and the heading error for each method.
Table~\ref{tab_realworld_result} reports the APE and maximum deviation, showing consistent multi-meter localization errors across all three SLAM systems (minimum APE: 3.46~m), which confirms real-world attack effectiveness.
Moreover, the observed drift in the real-world experiment was larger than in simulation.
We attribute this primarily to changing the LiDAR from the VLP-32c to the AT128.
Specifically, (i) the AT128 has higher angular resolution, making it more likely for beams to hit the mirror even from farther distances, and (ii) its field of view ($120^{\circ}\times25.4^{\circ}$) is narrower than the simulation setting ($180^{\circ}\times40.0^{\circ}$), which can relatively increase the fraction of mirror-induced ghost points within the overall observations.
These findings reveal that long-range, high-resolution LiDAR sensors, despite being designed for high performance, are highly vulnerable to specular-reflection-based attacks.

\begin{table}[tb]
\centering
\caption{Real-world experiment results.}
\vspace{-0.1in}
\label{tab_realworld_result}
\begin{tabular}{c|ccc}
\toprule
Metric & KISS-ICP~\cite{vizzo2023ral} & FAST-LIO2~\cite{fastlio2} & GLIM~\cite{koide2024glim} \\ \midrule
APE [m] & $6.028$ & $3.458$ & $4.772$  \\
Head. Error [deg] & $10.731$ & $7.264$ & $6.923$ \\
\bottomrule
\end{tabular}
\end{table}

\section{DISCUSSIONS}

\noindent\textbf{Limitations:} 
While our experiments successfully demonstrated the attack's effectiveness, practical limitations remain.
First, it requires proximity to the target trajectory: as shown in \S~\ref{eval_robustness_validation}, when the nearest distance exceeds 3.0\,m, the injected ghost points diminish and the benefit of optimized placement largely disappears. Such close-range placement (e.g., below 2.5\,m) is feasible in practical scenarios such as narrow roads, and therefore remains a realistic threat; moreover, even without the optimization advantage, occasional large deviations can still arise, which may translate into a meaningful safety risk in high-traffic environments.
Second, although the loose speed requirement for yaw oscillation allows for manual actuation (\S~\ref{eval_real_world_study}), automating the attack outdoors poses physical challenges. The mirror is highly susceptible to weather-induced disturbances (e.g., wind), which complicates the stable, long-term deployment of a motorized device.

\noindent\textbf{Countermeasures:}
Our results demonstrate that even with modern secure LiDAR sensors, additional defenses are strictly required against deliberate mirror exploitation.
One promising direction is spatial and multi-modal redundancy. Since relying on a single LiDAR is inherently vulnerable, integrating multiple LiDARs or sensors less sensitive to specular reflections (e.g., GNSS, wheel odometry) increases unaffected observations and directly suppresses registration errors.
Furthermore, while detecting and filtering mirror-induced ghost points~\cite{yang2008dealing, yang2010solving, yun2018reflection, lee2023learning} improves robustness, existing methods typically assume static mirrors; their effectiveness against dynamically manipulated mirrors remains an open challenge.

\section{Conclusion}
In this paper, we proposed \textit{MirrorDrift}, a physical attack that actively disrupts LiDAR-SLAM localization by generating mirror-induced ghost points. To sustain false correspondences during scan matching, we optimize the placement of a planar mirror and periodically oscillate its yaw angle.

In real-world experiments, we demonstrated that the proposed attack can succeed even against a secure LiDAR equipped with interference-mitigation mechanisms, for which conventional point-cloud injection attacks are difficult to apply. Across three widely used LiDAR-SLAM algorithms, the attack induced at least $3.4$\,m localization error, causing severe trajectory deviations that pose a critical safety risk

Overall, this study reveals that an adversary can inflict critical impact on LiDAR-SLAM by actively leveraging mirrors, and provides important insights toward designing more robust localization systems.

\bibliographystyle{IEEEtran}
\bibliography{citation.bib}

@inproceedings{sato2021dirty,
  title={Dirty road can attack: Security of deep learning based automated lane centering under $\{$Physical-World$\}$ attack},
  author={Sato, Takami and Shen, Junjie and Wang, Ningfei and Jia, Yunhan and Lin, Xue and Chen, Qi Alfred},
  booktitle={30th USENIX security symposium (USENIX Security 21)},
  pages={3309--3326},
  year={2021}
}

@inproceedings{sato2024lidar,
  title={LiDAR Spoofing Meets the New-Gen: Capability Improvements, Broken Assumptions, and New Attack Strategies},
  author={Sato, Takami and Hayakawa, Yuki and Suzuki, Ryo and Shiiki, Yohsuke and Yoshioka, Kentaro and Chen, Qi Alfred},
  booktitle={ISOC Network and Distributed System Security Symposium (NDSS)},
  year={2024},
  organization={ISOC}
}

@article{fukunagarandom,
  title={Random Spoofing Attack against LiDAR-Based Scan Matching SLAM},
  author={Fukunaga, Masashi and Sugawara, Takeshi},
  journal={VehicleSec2024},
  year={2024}
}

@inproceedings{nagata2025slamspoof,
  title={Slamspoof: Practical lidar spoofing attacks on localization systems guided by scan matching vulnerability analysis},
  author={Nagata, Rokuto and Koide, Kenji and Hayakawa, Yuki and Suzuki, Ryo and Ikeda, Kazuma and Sako, Ozora and Chen, Qi Alfred and Sato, Takami and Yoshioka, Kentaro},
  booktitle={2025 IEEE International Conference on Robotics and Automation (ICRA)},
  pages={15181--15187},
  year={2025},
  organization={IEEE}
}

@article{vizzo2023ral,
  author    = {Vizzo, Ignacio and Guadagnino, Tiziano and Mersch, Benedikt and Wiesmann, Louis and Behley, Jens and Stachniss, Cyrill},
  title     = {{KISS-ICP: In Defense of Point-to-Point ICP -- Simple, Accurate, and Robust Registration If Done the Right Way}},
  journal   = {IEEE Robotics and Automation Letters (RA-L)},
  pages     = {1029--1036},
  doi       = {10.1109/LRA.2023.3236571},
  volume    = {8},
  number    = {2},
  year      = {2023},
  codeurl   = {https://github.com/PRBonn/kiss-icp},
}

@article{fastlio2,
  author={Xu, Wei and Cai, Yixi and He, Dongjiao and Lin, Jiarong and Zhang, Fu},
  journal={IEEE Transactions on Robotics}, 
  title={FAST-LIO2: Fast Direct LiDAR-Inertial Odometry}, 
  year={2022},
  volume={38},
  number={4},
  pages={2053-2073},
  doi={10.1109/TRO.2022.3141876}
}

@article{koide2024glim,
  title={GLIM: 3D range-inertial localization and mapping with GPU-accelerated scan matching factors},
  author={Koide, Kenji and Yokozuka, Masashi and Oishi, Shuji and Banno, Atsuhiko},
  journal={Robotics and Autonomous Systems},
  volume={179},
  pages={104750},
  year={2024},
  publisher={Elsevier}
}

@article{yahia2025seeing,
  title={Seeing is Deceiving: Mirror-Based LiDAR Spoofing for Autonomous Vehicle Deception},
  author={Yahia, Selma and Alla, Ildi and Mohan, Girija Bangalore and Rau, Daniel and Singh, Mridula and Loscri, Valeria},
  journal={arXiv preprint arXiv:2509.17253},
  year={2025}
}

@inproceedings{fan2017point,
  title={A point set generation network for 3d object reconstruction from a single image},
  author={Fan, Haoqiang and Su, Hao and Guibas, Leonidas J},
  booktitle={Proceedings of the IEEE conference on computer vision and pattern recognition},
  pages={605--613},
  year={2017}
}

@inproceedings{akiba2019optuna,
  title={Optuna: A next-generation hyperparameter optimization framework},
  author={Akiba, Takuya and Sano, Shotaro and Yanase, Toshihiko and Ohta, Takeru and Koyama, Masanori},
  booktitle={Proceedings of the 25th ACM SIGKDD international conference on knowledge discovery \& data mining},
  pages={2623--2631},
  year={2019}
}

@inproceedings{yang2008dealing,
  title={Dealing with laser scanner failure: Mirrors and windows},
  author={Yang, Shao-Wen and Wang, Chieh-Chih},
  booktitle={2008 IEEE International Conference on Robotics and Automation},
  pages={3009--3015},
  year={2008},
  organization={IEEE}
}

@INPROCEEDINGS{10164614,
  author={Mo, Qian and Zhou, Yuhuai and Zhao, Xiaolei and Quan, Xinglin and Chen, Yihua},
  booktitle={2023 6th International Symposium on Autonomous Systems (ISAS)}, 
  title={A Survey on Recent Reflective Detection Methods in Simultaneous Localization and Mapping for Robot Applications}, 
  year={2023},
  volume={},
  number={},
  pages={1-6},
  doi={10.1109/ISAS59543.2023.10164614}}

@article{liu2022integrated,
  title={An integrated lidar-slam system for complex environment with noisy point clouds},
  author={Liu, Kangcheng},
  journal={arXiv preprint arXiv:2212.05705},
  year={2022}
}

@inproceedings{kobayashi2025invisible,
  title={Invisible but Detected: Physical Adversarial Shadow Attack and Defense on $\{$LiDAR$\}$ Object Detection},
  author={Kobayashi, Ryunosuke and Nomoto, Kazuki and Tanaka, Yuna and Tsuruoka, Go and Mori, Tatsuya},
  booktitle={34th USENIX Security Symposium (USENIX Security 25)},
  pages={7369--7386},
  year={2025}
}

@inproceedings{zhu2021can,
  title={Can we use arbitrary objects to attack lidar perception in autonomous driving?},
  author={Zhu, Yi and Miao, Chenglin and Zheng, Tianhang and Hajiaghajani, Foad and Su, Lu and Qiao, Chunming},
  booktitle={Proceedings of the 2021 ACM SIGSAC Conference on Computer and Communications Security},
  pages={1945--1960},
  year={2021}
}

@inproceedings{koch2015detection,
  title={Detection of specular reflections in range measurements for faultless robotic slam},
  author={Koch, Rainer and May, Stefan and Koch, Philipp and K{\"u}hn, Markus and N{\"u}chter, Andreas},
  booktitle={Robot 2015: Second Iberian Robotics Conference: Advances in Robotics, Volume 1},
  pages={133--145},
  year={2015},
  organization={Springer}
}

@article{tibebu2021lidar,
  title={Lidar-based glass detection for improved occupancy grid mapping},
  author={Tibebu, Haileleol and Roche, Jamie and De Silva, Varuna and Kondoz, Ahmet},
  journal={Sensors},
  volume={21},
  number={7},
  pages={2263},
  year={2021},
  publisher={MDPI}
}

@inproceedings{segal2009generalized,
  title={Generalized-icp.},
  author={Segal, Aleksandr and Haehnel, Dirk and Thrun, Sebastian and others},
  booktitle={Robotics: science and systems},
  volume={2},
  number={4},
  pages={435},
  year={2009},
  organization={Seattle, WA}
}

@inproceedings{yun2018reflection,
  title={Reflection removal for large-scale 3D point clouds},
  author={Yun, Jae-Seong and Sim, Jae-Young},
  booktitle={Proceedings of the IEEE Conference on Computer Vision and Pattern Recognition},
  pages={4597--4605},
  year={2018}
}

@article{yang2010solving,
  title={On solving mirror reflection in lidar sensing},
  author={Yang, Shao-Wen and Wang, Chieh-Chih},
  journal={IEEE/ASME Transactions on Mechatronics},
  volume={16},
  number={2},
  pages={255--265},
  year={2010},
  publisher={IEEE}
}

@article{lee2023learning,
  title={Learning-based reflection-aware virtual point removal for large-scale 3D point clouds},
  author={Lee, Oggyu and Joo, Kyungdon and Sim, Jae-Young},
  journal={IEEE Robotics and Automation Letters},
  volume={8},
  number={12},
  pages={8510--8517},
  year={2023},
  publisher={IEEE}
}

@inproceedings{sturm2012benchmark,
  title={A benchmark for the evaluation of RGB-D SLAM systems},
  author={Sturm, J{\"u}rgen and Engelhard, Nikolas and Endres, Felix and Burgard, Wolfram and Cremers, Daniel},
  booktitle={2012 IEEE/RSJ international conference on intelligent robots and systems},
  pages={573--580},
  year={2012},
  organization={IEEE}
}
\end{document}